\documentclass[10pt,twocolumn,letterpaper]{article}

\usepackage{comment}

 \usepackage[pagenumbers]{cvpr} 

\usepackage{algorithm}
\usepackage{algpseudocode}
\usepackage{amsmath} 
\usepackage{algorithm}
\usepackage{algorithmicx}
\usepackage{algpseudocode}
\usepackage{amsmath}
\usepackage{amssymb}
\usepackage{booktabs}      
\usepackage{amssymb}       
\usepackage{amsmath}       
\usepackage{graphicx}      
\usepackage{multirow}   
\usepackage{pifont}
\newcommand{\xmark}{\ding{55}}
\usepackage[abs]{overpic}
\usepackage{bm}

\definecolor{cvprblue}{rgb}{0.21,0.49,0.74}
\usepackage[pagebackref,breaklinks,colorlinks,allcolors=cvprblue]{hyperref}

\def\confName{CVPR}
\def\confYear{2025}

\title{FPAN: Mitigating Replication in Diffusion Models through the \underline{F}ine-Grained \underline{P}robabilistic \underline{A}ddition of \underline{N}oise to Token Embeddings}

\author{
Jingqi Xu\thanks{Equal contribution.} \quad 
Chenghao Li\footnotemark[1] \quad 
Yuke Zhang \quad 
Peter A. Beerel \\
University of Southern California \\
{\tt\small \{jingqixu, cli78217, yukezhan, pabeerel\}@usc.edu}
}

\begin{document}
\maketitle

\begin{abstract}
Diffusion models have demonstrated remarkable potential in generating high-quality images. However, their tendency to replicate training data raises serious privacy concerns, particularly when the training datasets contain sensitive or private information. Existing mitigation strategies primarily focus on reducing image duplication, modifying the cross-attention mechanism, and altering the denoising backbone architecture of diffusion models. 
Moreover, recent work has shown that adding a consistent 
small amount of noise to text embeddings can reduce replication to some degree. 
In this work, we begin by analyzing the impact of adding varying amounts of noise. Based on our analysis, we propose a fine-grained noise injection technique that probabilistically adds a larger amount of noise to token embeddings. We refer to our method as \textbf{Fine-grained Probabilistic Addition of Noise (FPAN)}.
Through our extensive experiments, we show that our proposed FPAN
can reduce replication by an average of 28.78\% compared to the baseline diffusion model without significantly impacting image quality, and outperforms the 
prior consistent-magnitude-noise-addition approach by 26.51\%.
Moreover, when combined with other existing mitigation methods, our FPAN approach can further reduce replication by up to 16.82\% with similar, if not improved, image quality.
\end{abstract}
    
\section{Introduction}
\label{sec:intro}

Diffusion models~\cite{dhariwal2021diffusion, ho2020denoising, rombach2022high} have become a dominant paradigm in generative modeling due to their strong capabilities in producing high-quality and diverse images. Compared to traditional approaches like Variational Autoencoders (VAEs)~\cite{kingma2019introduction} and Generative Adversarial Networks (GANs)~\cite{goodfellow2020generative}, diffusion models offer superior fidelity, diversity, and controllability. In particular, text-to-image diffusion models such as DALL·E~\cite{ramesh2021zero}, Stable Diffusion~\cite{rombach2022high}, and Imagen~\cite{saharia2022photorealistic} excel at generating images that are both semantically aligned with input captions and photorealistically detailed. These models iteratively denoise Gaussian noise based on textual prompts, producing aligned outputs after a fixed number of steps. Despite their success, recent studies~\cite{carlini2023extracting, somepalli2023diffusion, somepalli2023understanding} have shown that diffusion models are susceptible to memorizing training data, often generating outputs that closely resemble specific training images. This replication raises concerns over copyright infringement and privacy leakage, especially when models are fine-tuned on custom or sensitive small datasets~\cite{somepalli2023diffusion}.

Prior mitigation strategies~\cite{chen2024towards, gu2023memorization, li2024mitigate, schramowski2023safe, somepalli2023understanding, webster2023duplication, wen2024detecting, li2024loyaldiffusion, ren2024unveiling} that address the replication issue in diffusion models fall into three categories: 
optimization of the input image or text embeddings during training, modification of the cross-attention module, and architectural changes to the denoising backbone model. In the category of optimization of the input image or text embeddings, Somepalli et al.~\cite{somepalli2023understanding} introduce Random Token Replacement and Addition (RT), which randomly replaces tokens or inserts additional tokens into captions at random positions. Li et al. \cite{li2024mitigate} introduce the Dual Fusion method (DF), which leverages large language models (LLMs) to generalize captions and further mitigates replication by weighted fusing fine-tuning data with data from another source.
Modifications of the cross-attention module try to prevent diffusion models from overemphasizing tokens that are likely to lead to replication via masking \cite{ren2024unveiling}. Within the category of architectural modifications, Li et al.~\cite{li2024loyaldiffusion} improve the U-Net~\cite{ronneberger2015u} architecture by dynamically modifying the skip connections at specific timesteps to limit the impact of the replication-causing direct connections between the upsampling and downsampling blocks.

Prior work shows that perturbing text embedding with a small amount of noise provides a straightforward mitigation strategy \cite{somepalli2023understanding}, as specific captions have been shown to contribute to replication in diffusion models~\cite{li2024mitigate, somepalli2023understanding}. 
To enable finer-grained control over noise addition, and to explore the potential of increasing noise intensity for stronger replication mitigation,
we studied the impact of a much wider range of noise on token embeddings, and found that as noise intensity increases, the quality of the generated images first decreases, then improves, and eventually degrades again. Using a modified CLIPScore~\cite{pascual2024enhancing} to measure the degree of model overfitting, we show that this trend can be attributed to the model overfitting, well-fitting, and underfitting the training data, respectively.

Building upon this insight, we propose a \textbf{Fine-grained Probabilistic Addition of Noise (FPAN)} to  balance the trade-off between generation quality and replication. 
Our fine-tuning strategy operates at the fine-grained token embedding level and probabilistically injects relatively high-intensity noise into the tokens. Our empirical results demonstrate that our proposed method, in comparison to the baseline model, can significantly reduce replication score while preserving high generated image quality. When integrated with prior replication mitigation techniques~\cite{somepalli2023understanding, li2024loyaldiffusion, ren2024unveiling, li2024mitigate}, our method consistently enhances their performance, highlighting its potential for achieving synergistic improvements in diffusion model training.


We summarize our contributions as follows.
 1) We observe that as we increase the intensity of injected noise to token embedding the image quality tends to first becoming worse, then better, and then worse again and we show that this trend can be attributed to the model being in the states of overfitting, well-fitting, and underfitting, respectively. 2) We propose our FPAN strategy, which probabilistically injects high-intensity noise into fine-grained token embeddings during training. 3) We present experimental results demonstrating that our strategy provides competitive trade-offs between generation quality and replication. 4) We further show that our method can be effectively integrated with other mitigation techniques to achieve significant synergistic effects in further
 reducing replication.

\section{Background and Related works}
\label{sec:formatting}

\subsection{Diffusion Models}
Denoising Diffusion Probabilistic Models (DDPMs) \cite{ho2020denoising} employ a forward process that gradually adds Gaussian noise to an image and a reverse process that reconstructs the original image by progressively removing noise.
Due to high computational overhead in DDPMs, Latent Diffusion Models (LDMs), such as Stable Diffusion (SD) \cite{rombach2022high}, have been explored. LDMs apply a conditional diffusion process to a compressed latent space transformed by a Variational Autoencoder (VAE) \cite{kingma2019introduction}. 

Fine-tuning a pretrained SD model leverages a dataset consisting of $N$ image-caption pairs, expressed as $\mathcal{D} = \{(x^{(i)}, y^{(i)})\}_{i=1}^N$, where $x^{(i)}$ denotes the $i^{th}$ image, and $y^{(i)}$ represents its corresponding caption. 
During the forward corruption process, each clean image $x$ is progressively corrupted through the incremental addition of Gaussian noise over $T$ discrete timesteps, resulting in pure Gaussian noise. The noisy representation of an image $x$ at timestep $t$ is denoted by $x_t$, and the noise introduced at this timestep is represented as 
$\epsilon_{t}$. In the backward denoising process, the model $M$
is fine-tuned to estimate the noise added at each timestep $t$, conditioning on the caption $y$. The predicted noise is subsequently removed from the noisy input, facilitating a step-wise reconstruction of the original clean image. Formally, the optimization objective during training is expressed as follows:
\begin{equation}
    \mathcal{J}(\theta) = \mathbb{E}_{t \in [1,T], \epsilon_t \sim \mathcal{N}(0,I)} \left[ \|\epsilon_t - M
    (x_t, t, e)\|^2_2 \right],
\end{equation}
where $e$ is the text embedding obtained by applying CLIP \cite{radford2021learning} text encoder to original caption $y$.



\subsection{Replication Score}
To evaluate the degree of replication in generated images, we leverage the Replication Score \cite{li2024mitigate, li2024loyaldiffusion, somepalli2023diffusion, somepalli2023understanding}, denoted as $R$. $R$ is defined as the 95th-percentile statistic of the image-level similarity score between the generated images and their nearest matches in the training set.

In other words, a top-1 similarity for every generated image $x_{gen}$ is computed as:
\begin{equation}
    Sim_{\text{Top1}}(x_{\text{gen}})=\max_{x_d \in \mathcal{D}} \textit{sim}\bigl( \phi(x_{\text{gen}}),\, \phi(x_d) \bigr),
\end{equation}
where $\phi$ extracts image embeddings using SSCD \cite{pizzi2022self} \footnote{Here we use pretrained sscd\_disc\_large model. It can be found and downloaded from https://github.com/facebookresearch/sscd-copy-detection/tree/main}, and $\textit{sim}$ is typically dot product. Then $Sim_{\text{Top1}}(x_{\text{gen}})$ among all $x_{\text{gen}}$ are collected into a set to compute the replication score $R$ as:
\begin{equation}
    R(\mathcal{G})=\mathcal{Q_{\text{0.95}}}\{Sim_{\text{Top1}}(x_{\text{gen}})|x_{\text{gen}} \in \mathcal{G} \},
\end{equation}
where $\mathcal{G}$ is the generated image set and $\mathcal{Q_{\text{0.95}}}$ means the 95th-percentile value in the set.

The reason why $R$ only focuses on top $5\%$ of generated images by similarity is to ensure the evaluation focuses on replicated samples rather than the entire dataset, which otherwise may be misleading because most generations may be nowhere near direct copies, but a small fraction of generated images can still be very close to a training image. By zooming in on the right-hand tail of similarity score distribution, $R$ captures the worst-case copying behavior. Specifically, a higher $R$ indicates a higher level of replication.

\subsection{Mitigation strategies}
Prior researches~\cite{li2024mitigate, somepalli2023understanding} suggest that replication is primarily driven by image duplication and highly specific captions. To address this issue, several mitigation strategies have been proposed, which can be broadly categorized into three classes: model input-based strategies, cross-attention-based strategies, and architectural modifications. Within the category of model input-based strategies,
Li et al.~\cite{li2024mitigate} proposed a generality score and leveraged a large language model (LLM) \cite{achiam2023gpt, touvron2023llama} to increase caption abstraction.
They also introduced a dual fusion technique that merges training images with external image-caption pairs to address duplication. 
Anti-Memorization Guidance (AMG) \cite{chen2024towards} employs despecification, deduplication, and dissimilarity guidance to mitigate replication. Multiple Captions(MC)~\cite{somepalli2023understanding} uses BLIP to generate 20 captions per image and randomly samples one during each fine-tuning iteration. Random Caption Replacement (RC)~\cite{somepalli2023understanding} uses random words to replace the caption of an image. Caption Word Repetition(CWR)~\cite{somepalli2023understanding} randomly choose a word from the given caption and insert it into a random location in the caption.
\par 
In addition, various cross-attention-based strategies have been proposed. Ren et al.~\cite{ren2024unveiling} adjusted attention scores to reduce reliance on "trigger tokens" during inference, while chen et al.~\cite{chen2024exploring} introduced the Bright Ending (BE) mask to lower dependence on final prompt tokens. Zhang et al.~\cite{zhang2024forget} reduce the influence of specific tokens through attention resteering, effectively suppressing the model’s reliance on memorized concepts. Hintersdorf et al.~\cite{hintersdorf2024finding} proposed identifying and removing neurons in cross-attention layers responsible for replication. 
\par 
Furthermore, architectural modifications have also been explored. 
Li et al.~\cite{li2024loyaldiffusion} proposed RAU-Net, incorporating an Information Transfer Block into U-Net’s skip connections to prevent direct transmission of high-resolution information. 
\par 
Our FPAN offers a mitigation approach without sacrificing generation quality by probabilistically adding appropriate high-intensity noise to fine-grained token embeddings. Because of its orthogonal operational mechanism that ensures non-interference with existing techniques, FPAN can be synergistically combined with prior mitigation strategies to further reduce their replication scores.
\section{Fine-Grained Probabilistic Addition of Noise (FPAN)}

\subsection{Adding noise to token embeddings}
Research has shown that specific captions, corresponding to specific text embeddigns, can cause diffusion models to replicate training images~\cite{li2024mitigate, somepalli2023understanding}. Since noise can reduce caption specificity~\cite{lee2021learning}, one straightforward approach to mitigate this issue is to inject noise into the text embeddings. In FPAN, we propose to inject noise at the token embedding level to enable finer-grained control over the semantic perturbation.

Let \( \psi = \{ \tau_i \}_{i=1}^{L} \) denote a text embedding consisting of \( L \) token embeddings, where each \( \tau_i \in \mathbb{R}^{1 \times d} \) represents the \( i \)-th token embedding in a \( d \)-dimensional space. 
The following noise injection process are applied during each training iteration, 
\begin{equation}\label{eq:token_embedding}
\tau_i' = \tau_i + \boldsymbol{\xi}_i,\ \boldsymbol{\xi}_i \sim W \cdot \mathcal{N}(\mathbf{0}, \mathbf{I}),
\end{equation}
where \( \boldsymbol{\xi}_i \) 
denotes a Gaussian noise, and 
$W \in \mathbb R_{+}$ 
controlling the noise intensity.  
Let \( \psi' = \{ \tau_i' \}_{i=1}^{L} \) denote the noisy text embedding, where \( \tau_i' \in \mathbb{R}^{1 \times d} \) represents the noisy embedding of the \( i \)-th token.

\par
\begin{figure}[t]
\centering

\begin{minipage}{0.39\textwidth}  
  \centering
  \begin{overpic}[width=\linewidth]{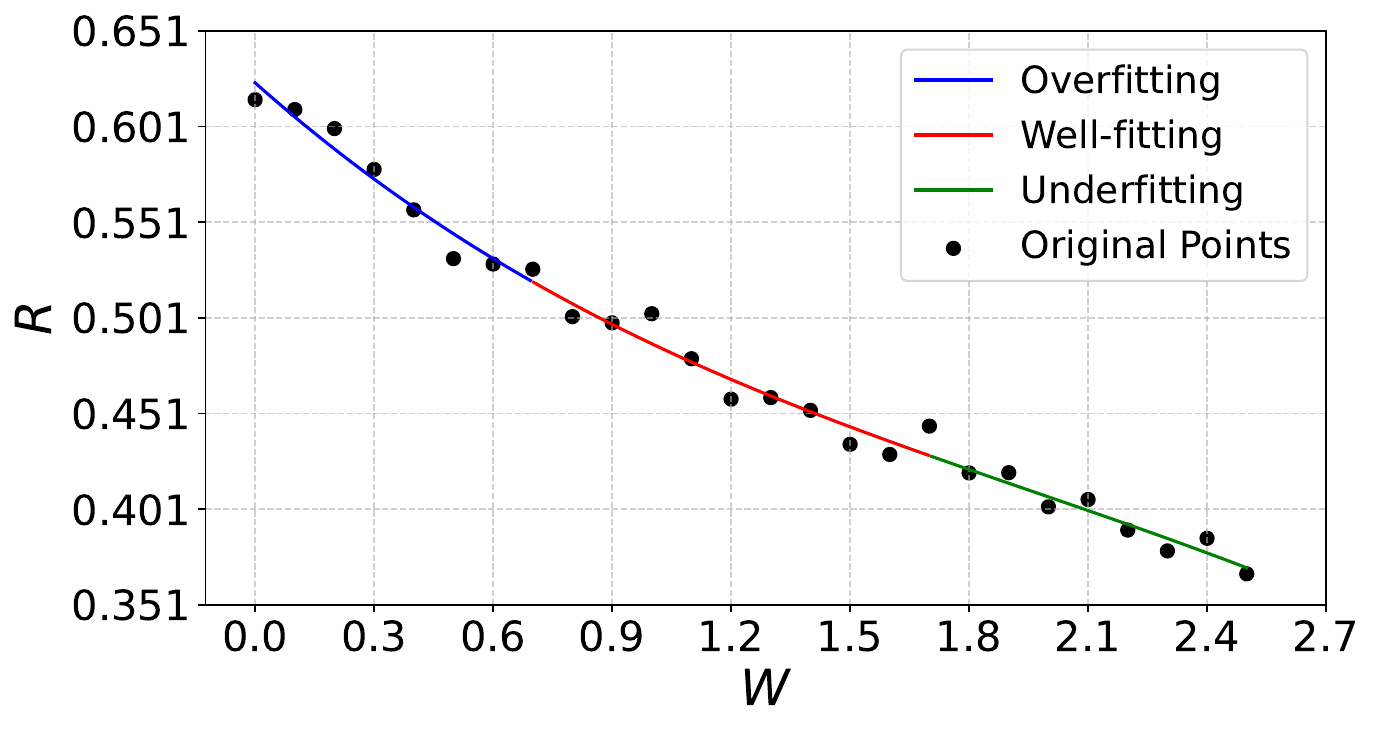}
    \put(2,3){\small (a)}  
  \end{overpic}
\end{minipage}

\vspace{0.03cm}

\begin{minipage}{0.39\textwidth}
  \centering
  \begin{overpic}[width=\linewidth]{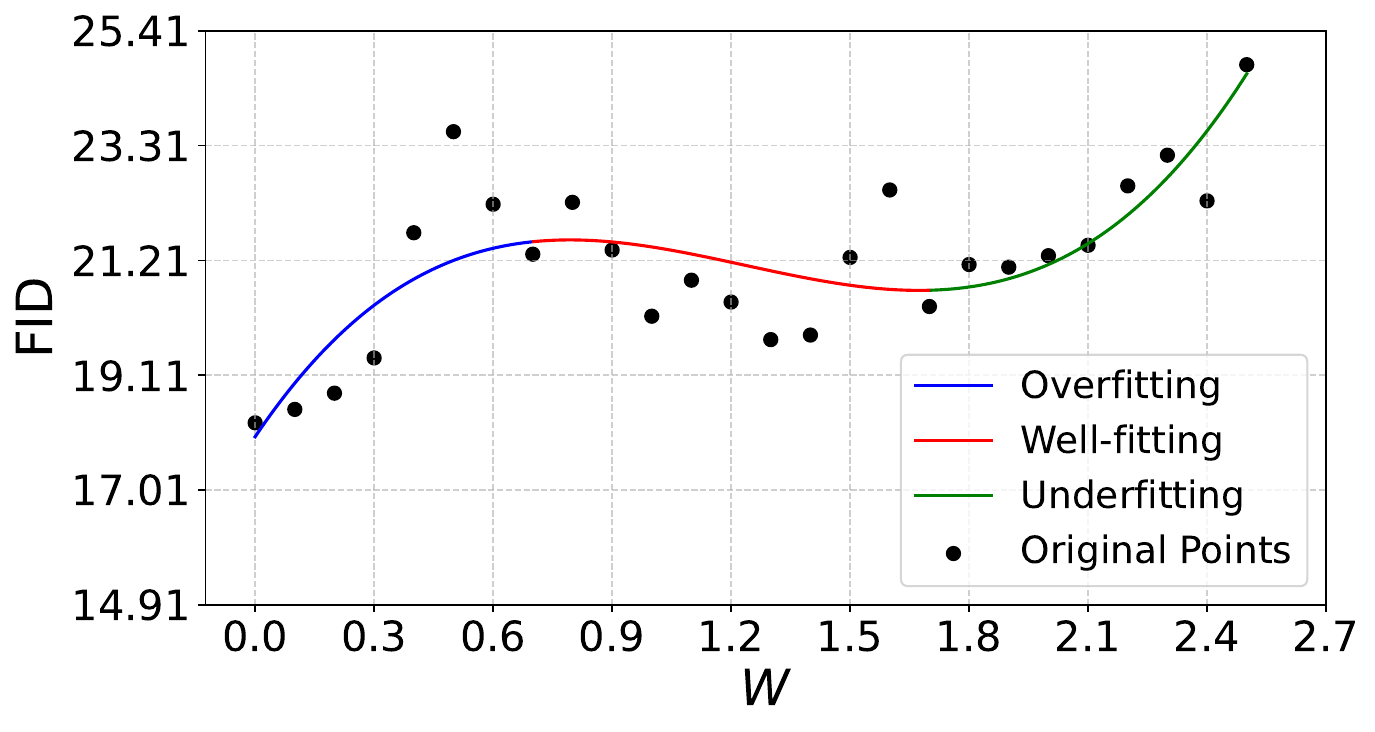}
    \put(2,3){\small (b)} 
  \end{overpic}
\end{minipage}

\vspace{0.03cm}

\begin{minipage}{0.39\textwidth}
  \centering
  \begin{overpic}[width=\linewidth]{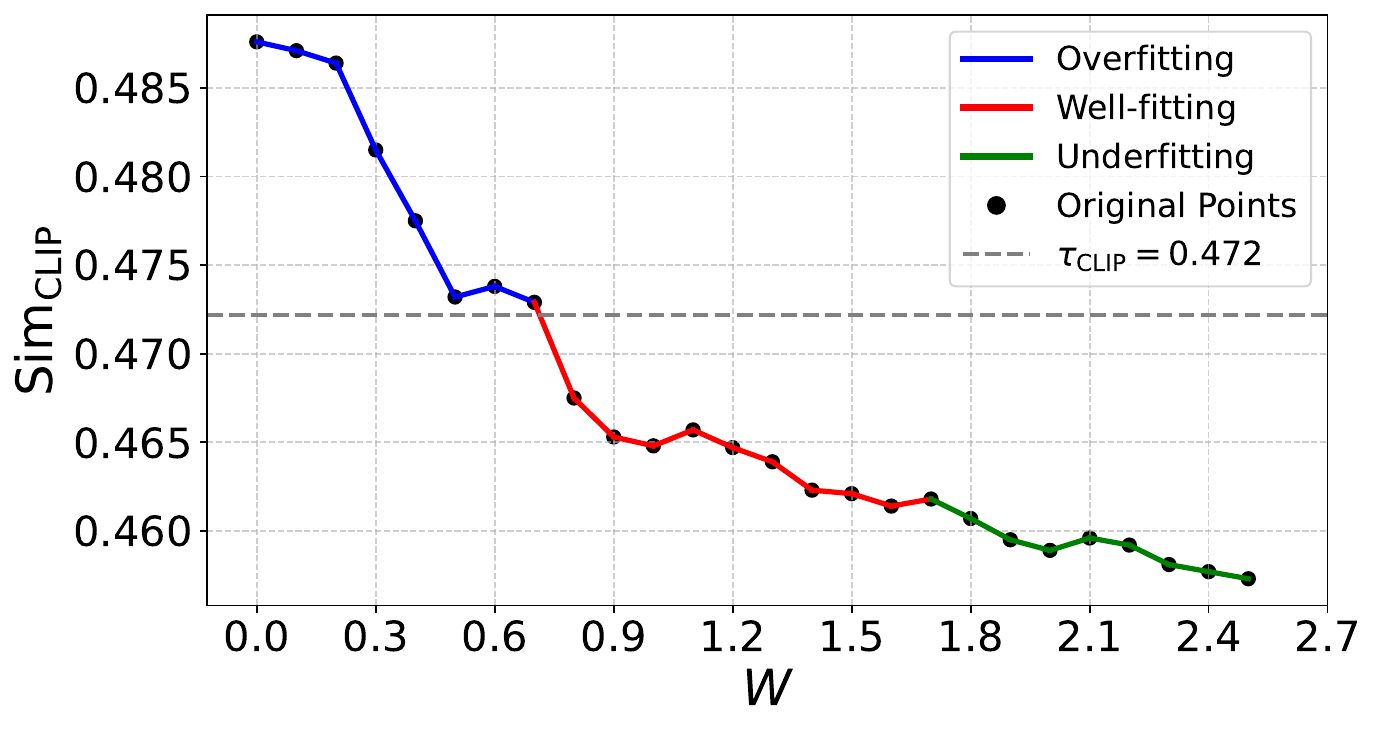}
    \put(2,3){\small (c)}  
  \end{overpic}
\end{minipage}

\vspace{-0.05cm}  

\caption{
Results under different noise intensities. (a) $R$, (b) FID, (c) \( \mathrm{Sim}_{\text{CLIP}} \). Three stages are shown as, overfitting stage (blue) when $W \leq 0.7$; well-fitting stage (red) when $0.7 < W \leq 1.7$; underfitting stage (green)when $W > 1.7$.}
\label{fig:three_metrics}
\end{figure}

\subsection{Analyzing impact of token noise intensity}
\label{section_3_2}

\begin{figure*}[t]
    \centering
    \includegraphics[width=0.65\textwidth]{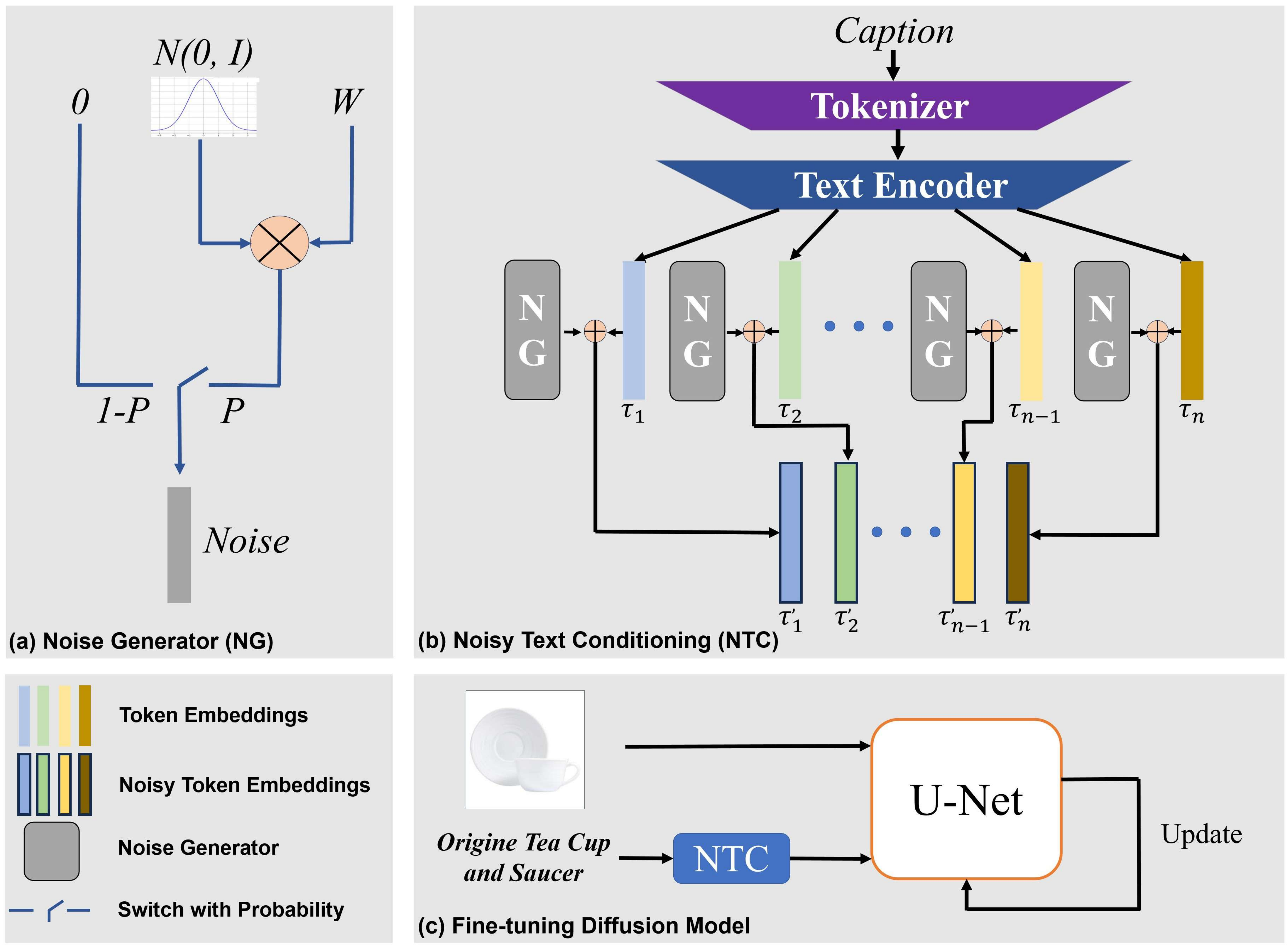}
    \caption{
    Overview of our proposed framework. (a) Probabilistic selection of noise. (b) Generation of noisy token embeddings. (c) Diffusion model fine-tuning process including noisy token embedding.
    }
    \label{fig:framework}
\end{figure*}

An important question that arises is how to determine the appropriate intensity $W$ of the added noise. Intuitively, adding noise with excessively high intensity may effectively distort the captions that are specific but can also degrade the sematic information in the caption and result in poor quality of generated images. Conversely, extremely low intensity noise may fail to effectively reduce the caption specificity and mitigate replication.

This section explores the impact of noise intensity by conducting experiments and comparing both replication and image quality across different values of $W$. Our experiments are based on Stable Diffusion 2.1~\cite{rombach2022high}. We fine-tune this diffusion model on a random subset of 10,000 samples from the LAION-2B dataset~\cite{schuhmann2022laion}, incorporating noise addition to the text embeddings as described in Equation~\ref{eq:token_embedding}. We vary $W$ from $0$ to $2.5$ in steps of $0.1$. To evaluate performance, we use the replication score $R$~\cite{somepalli2023diffusion, somepalli2023understanding, li2024mitigate, li2024loyaldiffusion} to quantify the replication, and the Fréchet Inception Distance (FID)~\cite{lucic2018gans} to assess the fidelity and diversity of the generated images. Additional details on the experimental setup are provided in Section~\ref{sec:experiment_setup}.

Our exprimental results are presented in Figure~\ref{fig:three_metrics}, where (a) illustrates how $R$ varies with the intensity of the injected noise by applying third-order polynomial fitting to the experimental data points. As \( W \) increases, the replication score $R$ monotonically decreases. 
This is consistent with prior work \cite{somepalli2023understanding} which argues that replication in text-to-image diffusion models is largely due to the over-learning of overly detailed semantic information in text embeddings and their associations with 
training images. As \( W \) increases, the level of semantic information contained in token embeddings is gradually 
degraded, making it more difficult to learn the relationship between text conditioning and their associated images,
leading to a gradual reduction in replication. As a result, $R$ exhibits a decreasing trend as $W$ increases.

The behavior of FID is less obvious as it shows a non-linear N-shaped increase–decrease–increase pattern as $W$ increases. 
As shown in Figure~\ref{fig:three_metrics} (b), we hypothesize that as $W$ increases, the model falls into one of three-stage, overfitting, well-fitting and finally underfitting. As $W$ increases in the overfitting stage, we see FID 
increases reflecting a reduction in overfitting, as adding noise to training data acts as a form of regularization that mitigates overfitting ~\cite{dhifallah2021inherent}. 
In the well-fitting stage, as \( W \) increases, FID begins to decline, a point that potentially be attributed to fully overcoming the overfitting stage and the ability to start to generalize. As \( W \) increases further, FID begins to rise again, which may be attributed to the model entering a stage of underfitting the training data.

To justify this hypothesis, we leverage a metric proposed in~\cite{pascual2024enhancing}, which we denote as \( \mathrm{Sim}_{\text{CLIP}} \). This metric measures the average CLIP~\cite{radford2021learning} embedding similarity between the generated image set and training image set. 
Specifically, for each generated image, we compute the average cosine similarity between its embedding and the embeddings of all images in the training dataset. The final \( \mathrm{Sim}_{\text{CLIP}} \) is then obtained by averaging these values across all generated images \cite{pascual2024enhancing}.  \( \mathrm{Sim}_{\text{CLIP}} \) serves as an indicator of overfitting, as higher values suggest that the generated images closely resemble those in the training set.  In particular, we define a threshold $\tau_{CLIP}$ to distinguish between well-generalized and overfitted models that is based on large pretrained models that are generally trained on diverse and extensive datasets and thus unlikely to overfit our small fine-tuning dataset~\cite{zeng2024infusion}. In particular, we generate a
reference image set with the pretrained model and our fine-tuning prompts and set
the threshold $\tau_{CLIP}$ to the \( \mathrm{Sim}_{\text{CLIP}} \) measured
between the reference and fine-tuning image sets. Fine-tuned models with \( \mathrm{Sim}_{\text{CLIP}} \) above $\tau_{CLIP}$ are flagged as potentially overfitting. 

In Figure~\ref{fig:three_metrics} (c), we present the \( \mathrm{Sim}_{\text{CLIP}} \) score 
for diffusion models fine-tuned with different noise intensities and 
compare them to $\tau_{CLIP} = 0.472$ measured with Stable Diffusion 2.1 \cite{rombach2022high}.
In the overfitting stage, we can see the \( \mathrm{Sim}_{\text{CLIP}} \) score is always larger than $\tau_{CLIP}$ and gradually decreases as \( W \) increases, supporting our hypothesis.
In the well-fitting stage, when \( W \) increases beyond 0.7, the \( \mathrm{Sim}_{\text{CLIP}} \) score drops below $\tau_{CLIP}$ and we assert our fine-tuned model no longer is overfitting. In the underfitting stage, excessive noise severely 
disrupts the semantic content of the captions, leading the model to generate 
significantly degraded images.

\par
\par
\par

\subsection{Probabilistic addition of noise}

In Section~\ref{section_3_2}, we show that the maximum noise intensity associated with the well-fitting stage will maximally mitigate replication without excessively degrading generation quality. However, compared to the case with no noise, the FID score can still be somewhat degraded. To 
compensate for the FID score increase and improve the replication-FID trade-off, we propose a fine-tuning process with Fine-grained Probabilistic Addition of Noise (FPAN) strategy. In particular, our method {\bf probabilistically} adds this maximal-intensity noise to each token embedding during each fine-tuning iteration. More precisely, instead of consistently adding high-intensity noise to all token embeddings, our probabilistic mechanism defines a probability factor \( P  \), which controls the probability of whether we inject high-intensity noise into a specific token embedding or not. 
The intensity $W$ of the injected noise is set to the maximum noise associated with 
the well-fitting stage because of its associated low $R$ and relative local minimum FID, as shown in Figures~\ref{fig:three_metrics} (a) and (b). 
An overview of the proposed method is given in Figure~\ref{fig:framework}.
More formally, let \( \psi = \{ \tau_i \}_{i=1}^{L} \) denote a text embedding consisting of \( L \) token embeddings, where each \( \tau_i \in \mathbb{R}^{1 \times d} \) represents the \( i \)-th token embedding in a \( d \)-dimensional space. During each fine-tuning iteration, we independently sample a noise term \( \boldsymbol{\xi}_i \in \mathbb{R}^{1 \times d} \) for each token embedding from the following distribution:
    
\begin{equation}
     \boldsymbol{\xi}_i \sim z_i \cdot \mathcal{N}(\mathbf{0}, W^2 \mathbf{I}), z_i \sim \text{Bernoulli}(P),
     \label{equation_5}
\end{equation}
where \( z_i \) is a Bernoulli random variable that equals to 1 with probability \( P \). Therefore, \( \boldsymbol{\xi}_i \) is sampled from \( \mathcal{N}(\mathbf{0}, W^2 \mathbf{I}) \) with probability \( P \), and is set to zero with probability \( 1 - P \). The noise \( \boldsymbol{\xi}_i \) is then injected into the corresponding token embedding as
\begin{equation}
    \tau_i' = \tau_i + \boldsymbol{\xi}_i,
\end{equation}
where \( \tau_i' \in \mathbb{R}^{1 \times d} \) denotes the \( i \)-th noisy token embedding. All \( \tau_i' \) are aggregated into a noisy text embedding \( \psi' = \{ \tau_i' \}_{i=1}^{L} \), which is used as the conditioning input for fine-tuning the diffusion model during the current fine-tuning iteration. More detailed pseudocode for 
this approach is presented in Algorithm~\ref{alg:token_noise}.

\begin{algorithm}[t]
\caption{Fine-tuning with FPAN}
\label{alg:token_noise}
\begin{algorithmic}[1]
\State \textbf{Input:} Pre-trained model \( M \); Fine-tuning dataset \( \mathcal{D} \); Total timesteps \( T \); Number of iterations \( N \); Image Encoder \( \mathcal{E}(\cdot) \); Text Encoder \( \text{CLIP}(\cdot) \); Noise weights \( W  \); Probability \( P  \).
\State \textbf{Output:} Fine-tuned model \( M \).
\State \( M.\text{train()} \)
\State \( \text{iter} \gets 0 \)
\While{\( \text{iter} < N \)}
    \For{each batch \( \{(x, y)\} \in \mathcal{D} \)}
        \State Image Encoding \( I \gets \mathcal{E}(x) \)
        \State Text Encoding \( \psi \gets \text{CLIP}(y) \)

        \For{each token embedding \( \tau_i \in \psi \)}
            \State Generate noise distribution:
            \State \( \Delta_{\mathit{}} \gets \mathcal{N}(0, W^2 \mathbf{I}) \)
            \State Sample Bernoulli variable: 
            \State \( z_i \sim \text{Bernoulli}(P) \)
            \State Sample noise: \( \boldsymbol{\xi}_i \gets z_i \cdot \Delta \)
            \State Add noise: \( \tau_i' \gets \tau_i + \boldsymbol{\xi}_i \)
        \EndFor

     \State Noisy text embedding: \( \psi' \gets \{ \tau_i' \} \)
    \State Sample \( t \sim \text{Uniform}(0, T) \)
    \State Update the model: \( M \gets \text{update}(M, I, \psi', t) \)
    \State \( \text{iter} \gets \text{iter} + 1 \)
    \EndFor
\EndWhile
\State \Return \( M \)
\end{algorithmic}
\end{algorithm}

\par
\par
\par

\section{Experimental Mitigation Results}
In this section, we perform a systematic tuning of the hyperparameters for our proposed method and provide analysis.
Furthermore, to evaluate the effectiveness of our approach, we present empirical evidence demonstrating its ability to mitigate replication, both as a standalone strategy and in combination with existing state-of-the-art methods.

\subsection{Experimental Setup} \label{sec:experiment_setup}

\begin{table*}[]
\centering
\resizebox{0.95\textwidth}{!}{
\begin{tabular}{c|ccccccccccr}
\hline  \hline
    \multicolumn{11}{c}{$W=1.5$}     \\ \hline
P      & 0     & 0.1     & 0.2     & 0.3     & 0.4   & 0.5    & 0.6    & 0.7    & 0.8    & 0.9    & 1.0    \\\hline
R$\downarrow$               & 0.615 & 0.559 & 0.538 & 0.512 & 0.491 & 0.477 & 0.474 & 0.461 & 0.457 & 0.451 &   0.434    \\\hline
FID$\downarrow$            & 18.24 & 17.37 & 16.43 & 17.45 & 17.77 & 18.15 & 18.91 & 19.87 & 18.22 & 18.83 &  21.26     \\ \hline \hline
 \multicolumn{11}{c}{$W=1.6$}   \\\hline
P      & 0     & 0.1     & 0.2     & 0.3     & 0.4   & 0.5    & 0.6    & 0.7    & 0.8    & 0.9    & 1.0    \\\hline
R$\downarrow$               & 0.615 & 0.559 & 0.525 & 0.488 & 0.475 & 0.469 & 0.457 & 0.450 & 0.452 & 0.440 & 0.429      \\\hline
FID$\downarrow$            & 18.24 & 17.26 & 16.35 & 17.07 & 17.21 & 17.44 & 17.83 & 20.44 & 19.31 & 22.27 &  22.49     \\ \hline \hline
 \multicolumn{11}{c}{$W=1.7$}  \\ \hline
P      & 0     & 0.1     & 0.2     & \textbf{0.3\textsubscript{\textbf{(A\textsubscript{1})}}}     & \textbf{0.4\textsubscript{\textbf{(A\textsubscript{2})}}}   & \textbf{0.5\textsubscript{\textbf{(A\textsubscript{3})}}}    & \textbf{0.6\textsubscript{\textbf{(A\textsubscript{4})}}}    & 0.7    & 0.8    & 0.9    & 1.0    \\\hline
R$\downarrow$               & 0.615 & 0.557 & 0.515 & \textbf{0.491} & \textbf{0.452} & \textbf{0.452} & \textbf{0.438} & 0.450 & 0.446 & 0.430 &  0.444     \\\hline
FID$\downarrow$            & 18.24 & 16.29 & 16.66 & \textbf{15.89} & \textbf{17.81} & \textbf{17.17} & \textbf{17.96} & 18.17 & 18.63 & 20.71 &  20.36     \\ \hline \hline
\end{tabular}}
\caption{Replication score and FID under different values of $P$ for three configurations: $W=1.5$, $W=1.6$, and $W=1.7$. For $W=1.7$, the values of $P=0.3$, $0.4$, $0.5$, and $0.6$ correspond to points A\textsubscript{1}, A\textsubscript{2}, A\textsubscript{3}, and A\textsubscript{4} in Figure~\ref{fig:R-FID1}, respectively.}
\label{tab:various-SC-result}
\end{table*}


\textbf{Model Selection and Dataset}
We build upon Stable Diffusion 2.1 \cite{rombach2022high}, an advanced text-to-image diffusion model pretrained on the complete LAION dataset \cite{schuhmann2022laion}. 
We fine-tune the model using a random subset of 10,000 samples from the LAION-2B dataset~\cite{schuhmann2022laion}. Each sample includes an image paired with a descriptive caption, thereby capturing a wide variety of visual and textual content.
The unmodified fine-tuned model serves as our baseline. Our study specifically targets adjustments to the noise addition strategy on text token embeddings during 
fine-tuning, leaving the model architecture intact.

\textbf{Fine-tuning process}
During fine-tuning, all components except the U-Net remain frozen. We adhered to the fine-tuning configuration detailed in~\cite{somepalli2023diffusion, somepalli2023understanding}, running the process for 100,000 steps with a learning rate of $5 \times 10^{-6}$ and incorporating a warm-up phase over the first 5,000 steps. The diffusion process is executed with $T=1000$ timesteps. For evaluation, we generate 10,000 images during the inference process using $50$ inference steps, with prompts identical to those used in the fine-tuning set. More implementation details are provided in Appendix~\ref{sec:experiment_setup}.

\textbf{Evaluation Metrics}
Our evaluation framework utilizes three key metrics. (1) Replication score ($R$)~\cite{somepalli2023diffusion, somepalli2023understanding, li2024loyaldiffusion, li2024mitigate}, which quantifies the degree of replication; (2) Frechet Inception Distance (FID)~\cite{lucic2018gans}, which assesses the fidelity and diversity of the generated images; and (3) the R-FID curve~\cite{li2024loyaldiffusion}, which illustrates the trade-off between replication and generation quality created by varying a shared hyperparameter across different methods. 
A second-order polynomial function is then fitted to the ($R$, FID) pairs to form a continuous trade-off curve. Curves that lie closer to the origin reflect more
favorable trade-offs.

\subsection{Hyperparameters Tuning}
\label{section4_2}
\begin{figure}[t]
    \centering
    \includegraphics[width=0.70\linewidth]{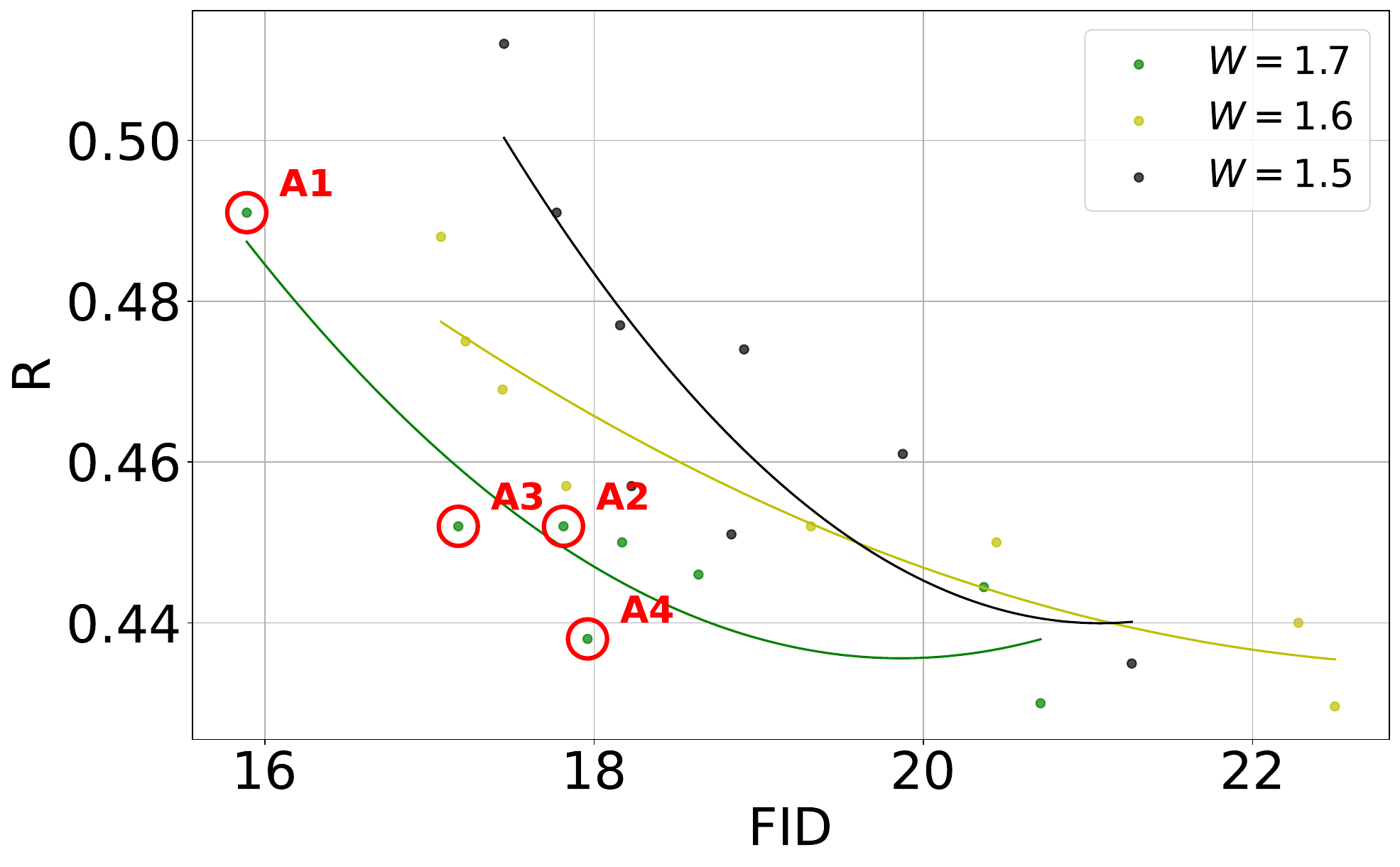}
    \caption{R-FID curves corresponding to different values of \( W \), including four favorable points corresponding to $W=1.7$. 
$A_1 (15.89, 0.491)$ when \( P = 0.3 \), $A_2 (17.81, 0.452)$ when \( P = 0.4 \),
$A_3 (17.17, 0.452)$ when \( P = 0.5 \), and $A_4 (17.96, 0.438)$ when \( P = 0.6 \).}
    \label{fig:R-FID1}
\end{figure}

Based on the observations from Figure~\ref{fig:three_metrics}, we consider three noise weight intensities: \( W = 1.5 \), \( W = 1.6 \), and \( W = 1.7 \), and vary the probability \( P \) uniformly over the interval \([0, 1.0]\) with a step size of \( 0.1 \).
To assess the effectiveness of different high-intensity noise levels, we leverage R-FID curves obtained by varying the probability parameter \( P \) for all three values of $W$.
Interestingly, we observed an interesting shift: as \( P \) decreases from 1 to approximately 0.2, the FID score decreases, but then rises sharply below this point. A detailed analysis of this behavior is provided in Appendix~\ref{sec:appendix_fid_trend}. For a clearer visual comparison of the R-FID curves, Figure~\ref{fig:R-FID1} presents curves fitted using results 
with \( 0.2 < P \leq 1 \).
The results show that the R-FID curve corresponding to \( W = 1.7 \) lies closest to the origin, indicating this value leads to the most favorable trade-off between image quality and replication mitigation. In Table~\ref{tab:various-SC-result}, we present the detailed results for all values of \( P \) under different settings of \( W \). It can be seen that when \( 0.3 \leq P \leq 0.6 \) under \( W = 1.7 \), replication is effectively mitigated while the corresponding FID scores remain lower than that of the baseline model. Specifically, the values \( P = 0.3 \), \( P = 0.4 \), \( P = 0.5 \), and \( P = 0.6 \) correspond to points A\textsubscript{1}, A\textsubscript{2}, A\textsubscript{3}, and A\textsubscript{4} in Figure~\ref{fig:R-FID1}, respectively. 

\subsection{Standalone Performance}

\begin{table*}[t]
\centering
\renewcommand{\arraystretch}{1.3}
\begin{tabular}{cccccccccc}
\hline\hline

& \multirow{2}{*}{Baseline} 
& GN
& MC
& RC
& CWR
& LD 
& DF
& TMAA
& FPAN(Ours) \\ 
& & ~\cite{somepalli2023understanding} & \cite{somepalli2023understanding} & \cite{somepalli2023understanding} &\cite{somepalli2023understanding} &\cite{li2024loyaldiffusion} &\cite{li2024mitigate}& \cite{ren2024unveiling} & (P\;=\;0.3\;/\;0.4\;/\;0.5\;/\;0.6) \\
\hline
$R \downarrow$   & 0.615 & 0.596 &0.420 & 0.565 & 0.614 & 0.378 & 0.412 & 0.309 & 0.491\;/\;0.452\;/\;0.452\;/\;0.438 \\
\hline
FID $\downarrow$ & 18.24 & 19.50 &16.83 & 15.98 & 16.73 & 19.17 & 17.47 & 38.18 & 15.89\;/\;17.81\;/\;17.17\;/\;17.96 \\
\hline\hline
\end{tabular}

\caption{Comparison of FPAN with prior works.}
\label{tab:method_compare}
\end{table*}

We employ our mitigation strategy with \( W = 1.7 \) under probability settings \( P \in \{0.3, 0.4, 0.5, 0.6\} \), and independently evaluate its effectiveness against a range of existing approaches. These include Gaussian Noise (GN)~\cite{somepalli2023understanding}, Multiple Captions (MC)~\cite{somepalli2023understanding}, Random Caption Replacement (RC)~\cite{somepalli2023understanding}, Caption Word Repetition (CWR)~\cite{somepalli2023understanding}, Loyal Diffusion (LD)~\cite{li2024loyaldiffusion}, Dual Fusion (DF)~\cite{li2024mitigate} and an inference-time method involving token masking and attention score adjustment (TMAA)~\cite{ren2024unveiling}. The corresponding results are summarized in Table~\ref{tab:method_compare}.

Compared to the baseline, our method reduces the replication score \( R \) by up to 28.78\%, while improving the FID score by up to 2.35. Compared to the GN, RC, and CWR methods, our approach achieves maximum reductions in \( R \) by 26.51\%, 22.47\%, and 28.66\%, respectively, without incurring significant degradation in the FID score. Moreover, compared to MC, LD, DF, and TMAA, our method yields improvements in FID by up to 0.94, 3.28, 1.58, and 22.29, respectively, while maintaining good R scores.

The above findings suggest that our method outperforms most 
other methods, when deployed as a standalone strategy, offers similar if not improved trade-off between memorization mitigation and generative quality over existing approaches.


\par

\subsection{Synergy with Prior Art}
To further demonstrate the benefits of our method, we investigate its synergistic effects when integrated with prior approaches that target replication mitigation from different perspectives. Specifically, we combine our strategy with four representative methods: Multiple Captions (MC)~\cite{somepalli2023understanding}, Dual Fusion (DF)~\cite{li2024mitigate}, LoyalDiffusion (LD)~\cite{li2024loyaldiffusion}, and Token Masking and Attention score Adjustment (TMAA)~\cite{ren2024unveiling}. 
The achieved best experimental results with corresponding \( P \) in our method are shown in Table~\ref{tab:with_without_nate}.
When combined with prior methods, our approach yields up to a 16.82\% reduction in \( R \).
In fact, our approach, when combined with LD yields the lowest $R = 0.357$ across all prior methods tested that have a reasonable $FID$ of less than $20$. The only lower achieved \( R \) is for TMAA with FPAN, where we get an $R =0.257$ but at the cost of a much higher $FID=36.93$. 

We may also note that the results show that in most cases, the addition of FPAN not only yields improvements in $R$ but also improvements in $FID$. The exception is MC with FPAN which shows a small increase in FID compared to MC alone. This may be attributed to the fact that the MC's multiple captions may already have sufficient randomness in the semantic information and adding more randomness using FPAN may be leading to some semantic degradation. However, the FID of MC with FPAN is still similar to baseline's FID, and thus it maintains high generation quality.

\begin{table}[t]
\centering
\renewcommand{\arraystretch}{1.3}
\begin{tabular}{c|p{1.4cm}|cc} 
\hline\hline
Method & w/FPAN & R$\downarrow$ & FID$\downarrow$ \\
\hline
\multirow{2}{*}{Baseline} & \hspace{0.3cm}\xmark     & 0.615 & 18.24 \\
                          & \hspace{0.3cm}\checkmark{\scriptsize (P=0.6)} & 0.438 & 17.96 \\
\hline
\multirow{2}{*}{MC~\cite{somepalli2023understanding}}       & \hspace{0.3cm}\xmark     & 0.420 & 16.83 \\
                          & \hspace{0.3cm}\checkmark{\scriptsize (P=0.4)} & 0.378 & 18.45 \\
\hline
\multirow{2}{*}{LD~\cite{li2024loyaldiffusion}}       & \hspace{0.3cm}\xmark     & 0.378 & 19.17 \\
                          & \hspace{0.3cm}\checkmark{\scriptsize (P=0.6)} & 0.357 & 19.78 \\
\hline
\multirow{2}{*}{TMAA~\cite{ren2024unveiling}}     & \hspace{0.3cm}\xmark     & 0.309 & 38.18 \\
                          & \hspace{0.3cm}\checkmark{\scriptsize (P=0.6)} & 0.257 & 36.93 \\
\hline
\multirow{2}{*}{DF~\cite{li2024mitigate}}       & \hspace{0.3cm}\xmark     & 0.412 & 17.47 \\
                          & \hspace{0.3cm}\checkmark{\scriptsize (P=0.6)} & 0.371 & 17.58 \\
\hline\hline
\end{tabular}
\caption{The impact of combining FPAN with prior works. \xmark\,
represents that the method is perform standalone and \checkmark\, means FPAN is also used with the method.}
\label{tab:with_without_nate}
\end{table}


\section{Ablation Studies}
\subsection{Fine-Grained vs. Coarse-Grained}
To justify the effectiveness of targeting fine-grained token embeddings, we compare FPAN against a coarser grained probabilistic addition of noise we refer to as CPAN.
In particular, while adopting the same sampling scheme for the noise term as described in Equation~\ref{equation_5}, CPAN applies the noise term to the entire text embedding rather than to individual token embeddings.
More specifically, in FPAN, each token has an independent probability of being perturbed by noise or left unchanged. However, in CPAN, if a text embedding is selected to be perturbed, all token embeddings within this text embedding are injected with noise. 
Otherwise, no token embedding within the text 
embedding will have injected noise.

\begin{figure}[t]
\centering

\begin{minipage}{0.35\textwidth}  
  \centering
  \begin{overpic}[width=\linewidth]{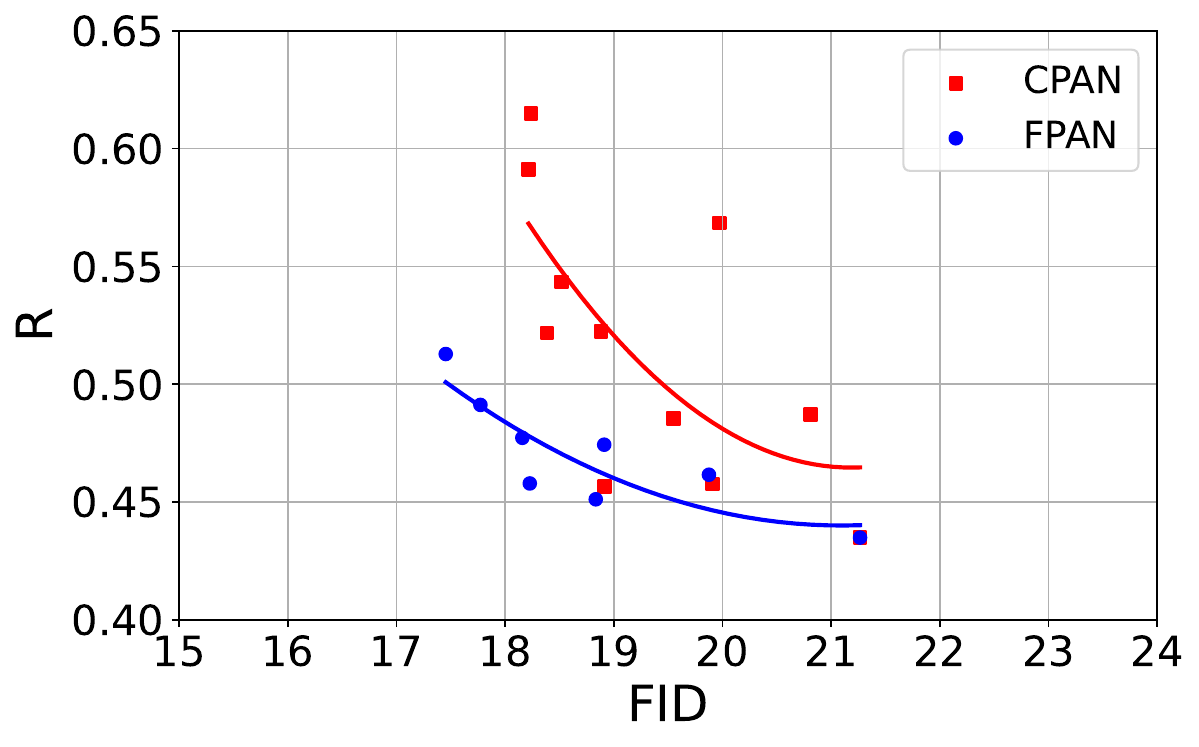}
    \put(2,-4){\small (a) W=1.5}  
  \end{overpic}
\end{minipage}
\vspace{0.3cm}  

\begin{minipage}{0.35\textwidth}
  \centering
  \begin{overpic}[width=\linewidth]{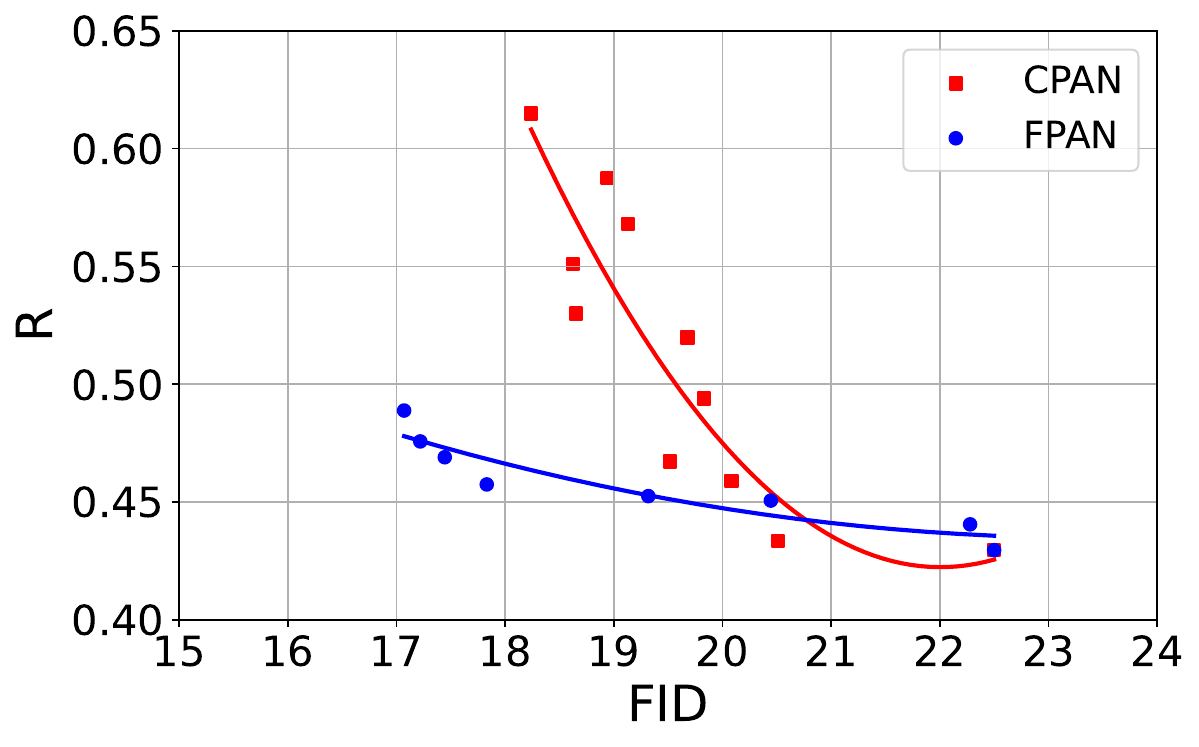}
    \put(2,-4){\small (b) W=1.6} 
  \end{overpic}
\end{minipage}
\vspace{0.3cm} 

\begin{minipage}{0.35\textwidth}
  \centering
  \begin{overpic}[width=\linewidth]{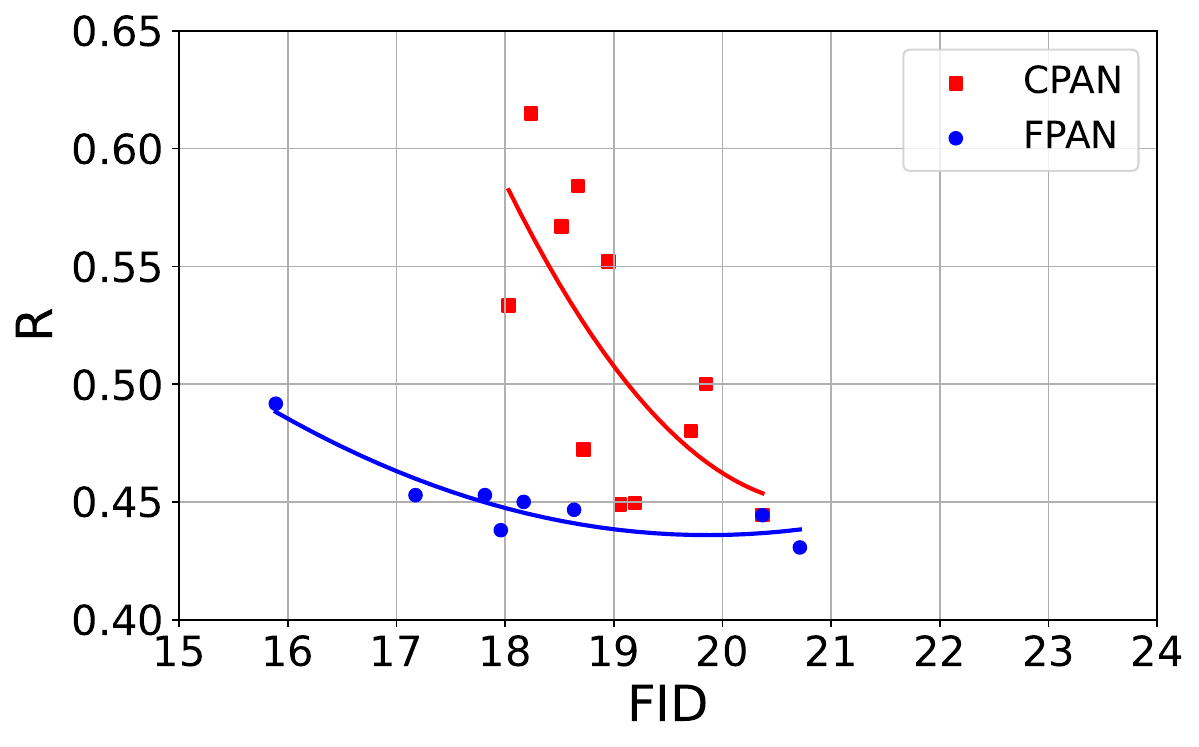}
    \put(2,-4){\small (c) W=1.7}  
  \end{overpic}
\end{minipage}

\vspace{0.2cm} 

\caption{
        Comparison of R-FID curves between FPAN and CPAN for various values of $W$. 
    }
\label{fig:FPAN_vs_CPAN}
\end{figure}

To compare the two approaches, we adopt the same hyperparameter settings as described in Section~\ref{section4_2}. For each possible value of \( W \), we observe R-FID curves by varying probability $P$ for both FPAN and CPAN. The results are presented in Figure~\ref{fig:FPAN_vs_CPAN}.
Across all values of \( W \), FPAN consistently produces R-FID curves that lie closer to the origin, indicating FPAN is more effective in mitigating $R$ while preserving FID.

We attribute the superior performance of our proposed FPAN approach to its more fine-grained and flexible noise injection strategy. Unlike CPAN, which often perturbs all token embeddings that can severely degrades semantic content, FPAN injects noise into individual token embeddings. 
This fine-grained process increases the chance of targeting specific tokens most responsible for replication, while leaving others intact. In principle, precisely controlling noise injection for specific tokens could offer even greater benefits, but doing so would require substantial computational overhead due to the need for token-level importance estimation or gradient-based analysis.


\subsection{Random Masking vs. FPAN}
\label{sec:RandomMaskingVsFPAN}
We refer to the approach in which each token embedding in a text embedding is masked with a probability \( Q \) during fine-tuning as the Random Masking (RM) strategy. We compare it with our FPAN approach because both of them aim to despecify text conditioning for diffusion models.

\begin{figure}[t]
    \centering
    \includegraphics[width=0.665\linewidth]{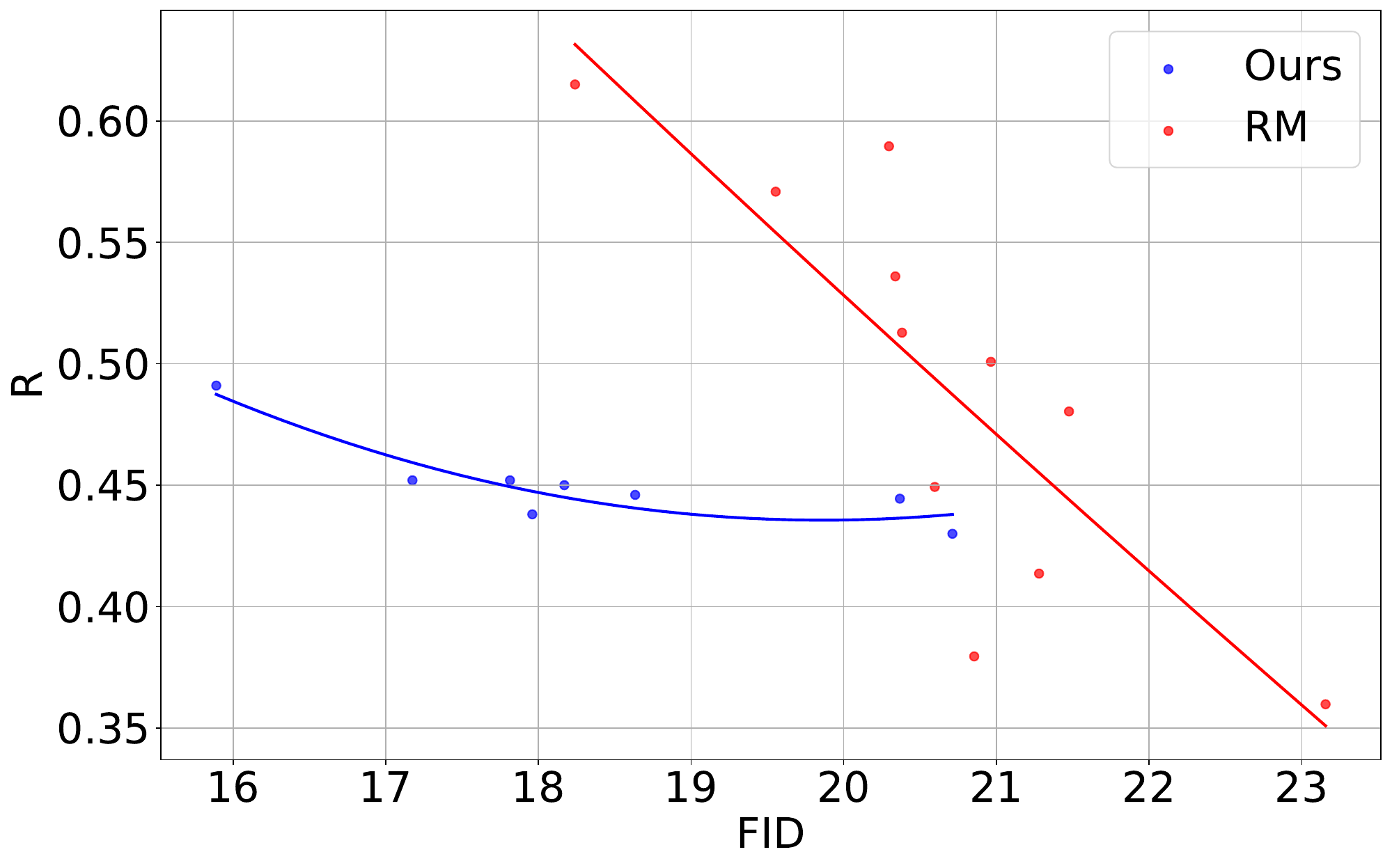}
    \caption{Comparison of R-FID curves between FPAN and RM.}
    \label{fig:ours_vs_rm_rfid}
\end{figure}

We obtain R-FID curves by varying the probability $P$ and $Q$ for both approaches, as shown in Figure~\ref{fig:ours_vs_rm_rfid}. Since RM does not exhibit the sudden shift phenomenon, its curve is fitted using results from \( 0 \leq Q \leq 1 \). For a clearer visual comparison, the curve corresponding to FPAN is fitted using the same \( P \) settings as described in Section~\ref{section4_2}, under the optimal hyperparameter setting of \( W = 1.7 \). 
The R-FID curve of FPAN lies closer to the origin 
and, in particular, shows FPAN outperforms RM in lower FID situations. In addition, more detailed analysis is  given in Appendix~\ref{appendix_3}

\section{Conclusions and Future Work}
In this work, we propose Fine-grained Probabilistic Addition of Noise (FPAN), a fine-tuning strategy designed to mitigate replication in diffusion models while maintaining generation quality. Our method probabilistically adds high-intensity noise to fine-grained token embeddings during each fine-tuning iteration. The choice of appropriate high-intensity noise is determined by our finding on how different amount of noise affect replication and generation quality. Through extensive experiments, FPAN demonstrates a significant reduction in replication compared to baseline models and prior works, without  compromising the FID score. Moreover, FPAN can be combined with recent mitigation methods to produce synergistic effects, further enhancing their performance.

While FPAN demonstrates strong performance, there are limitations to its current design. Specifically, both the noise intensity \( W \) and the probability parameter \( P \) are fixed throughout the training process. However, the optimal values of \( W \) and \( P \) may vary across different stages of training. Future research could explore dynamically adjusting \( W \) and \( P \) as training progresses, allowing more precise control over noise perturbation and further improving the trade-off between generation quality and replication mitigation.

\par


\clearpage
\setcounter{page}{1} 
\renewcommand{\thesection}{\Alph{section}} 
\setcounter{section}{0} 
\renewcommand{\thesubsection}{\Alph{section}.\arabic{subsection}} 
\maketitlesupplementary

\section{Appendix}
\label{sec:appendix}

\subsection{Additional Experimental Setup Details}
\label{sec:additional_experiment_setup}
All experiments are conducted using NVIDIA A100 Tensor Core GPUs equipped with 40\,\textit{GB} of memory, supporting both the training and inference process. During training, we set the batch size to 16 and fix the image resolution to 256. Optimization is performed using the Adam optimizer with \( \beta_1 = 0.9 \) and \( \beta_2 = 0.999 \), along with a weight decay factor of \( 1e^{-2} \). For the inference process, we generate samples using \( S = 50 \) steps, uniformly spacing across the full diffusion process.

\subsection{Analysis of the Sudden Shift in FID}
\label{sec:appendix_fid_trend}

\begin{figure}[t]
    \centering
    \includegraphics[width=0.48\textwidth]{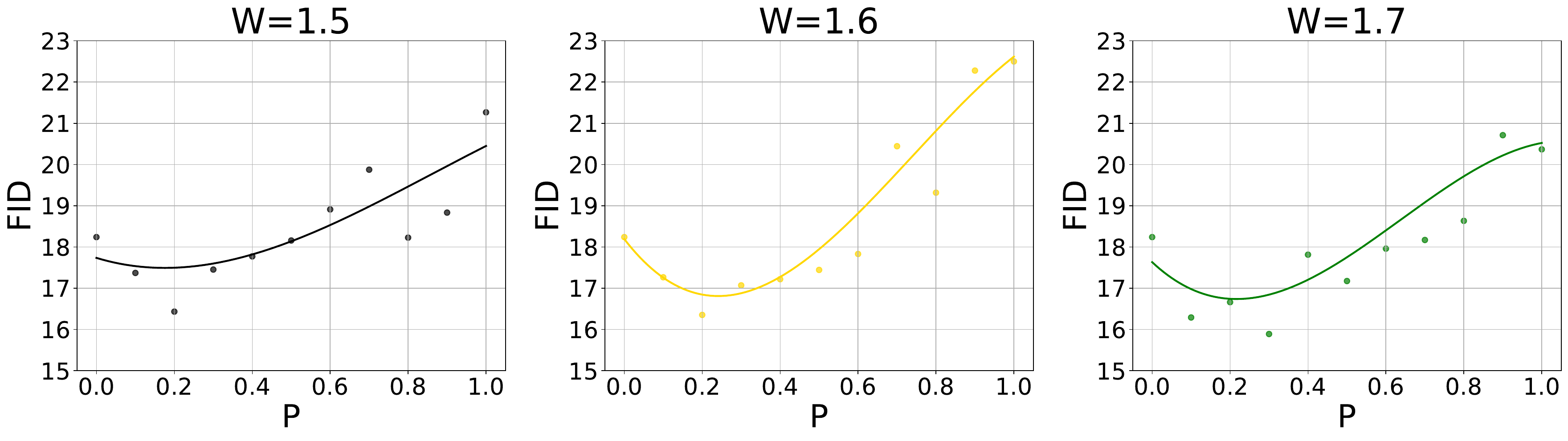}
    \caption{FID curve of FPAN as we vary \( P \) for three different values of $W = 1.5, 1.6, 1.7$.}
    \label{fig:fid_vs_p_avg_all_settings}
\end{figure}

In this section, we provide a detailed analysis of the nonlinear trend observed in the FID score as the probability parameter \( P \) decreases. In Figure~\ref{fig:fid_vs_p_avg_all_settings}, we present third-order polynomially fitted curves based on the FID scores obtained across different \( W \) for \( 0 \leq P \leq 1 \). We observe that the FID score steadily decreases as \( P \) decreases from 1 to approximately 0.2, but begins to rise abruptly below this point. 
We provide a possible explanation for this phenomenon.

According to~\cite{ure1971lexical}, approximately 40\% of the words in a sentence contribute significantly to its overall semantics, while the remaining 60\% are function words that carry relatively little semantic content. When \( P = 1 \),  all token embeddings that corresponding to high-information words are perturbed with strong noise, substantially disrupting the semantic integrity of the text embedding. Consequently, the quality of generated images is severely degraded, resulting in high FID scores. As \( P \) decreases, the proportion of high-information token embeddings subjected to strong noise gradually decreases. This leads to a progressive restoration of the global semantics encoded in the text embedding, thereby improving the generation quality and reducing the FID score. Furthermore, Clark et al.~\cite{clark2020electra} shows that when less than approximately 20\% of the words in a sentence are perturbed or replaced, a semantic classifier is still capable of identifying the key semantic content, indicating that the overall semantics are preserved. Building on this insight, when \( P \) decreases to approximately $0.2$, the proportion of token embeddings affected by strong noise reduces to around 20\%. 
This may explain the plateau observed in FID reduction: at this point, the original semantic information in the text embedding is almost fully recovered, and further decreases in \( P \) may no longer contribute to generation quality improvement. However, as \( P \) continues to decrease below $0.2$, the number of function word embeddings subjected to strong noise continues to decrease. Although function words individually encode limited semantic content, perturbing them introduces stochasticity that acts similarly to data augmentation~\cite{lee2021learning}, thereby enhancing the model’s generalization ability. 
We attribute the sudden increase in FID scores to the diminishing effectiveness of this regularization effect, which in turn may weaken the model's generalization capacity and lead to a decline in image generation quality.

\subsection{Comparison between FPAN and RM}
\label{appendix_3}

\begin{figure}[t]
\centering

\begin{minipage}{0.35\textwidth}  
  \centering
  \begin{overpic}[width=\linewidth]{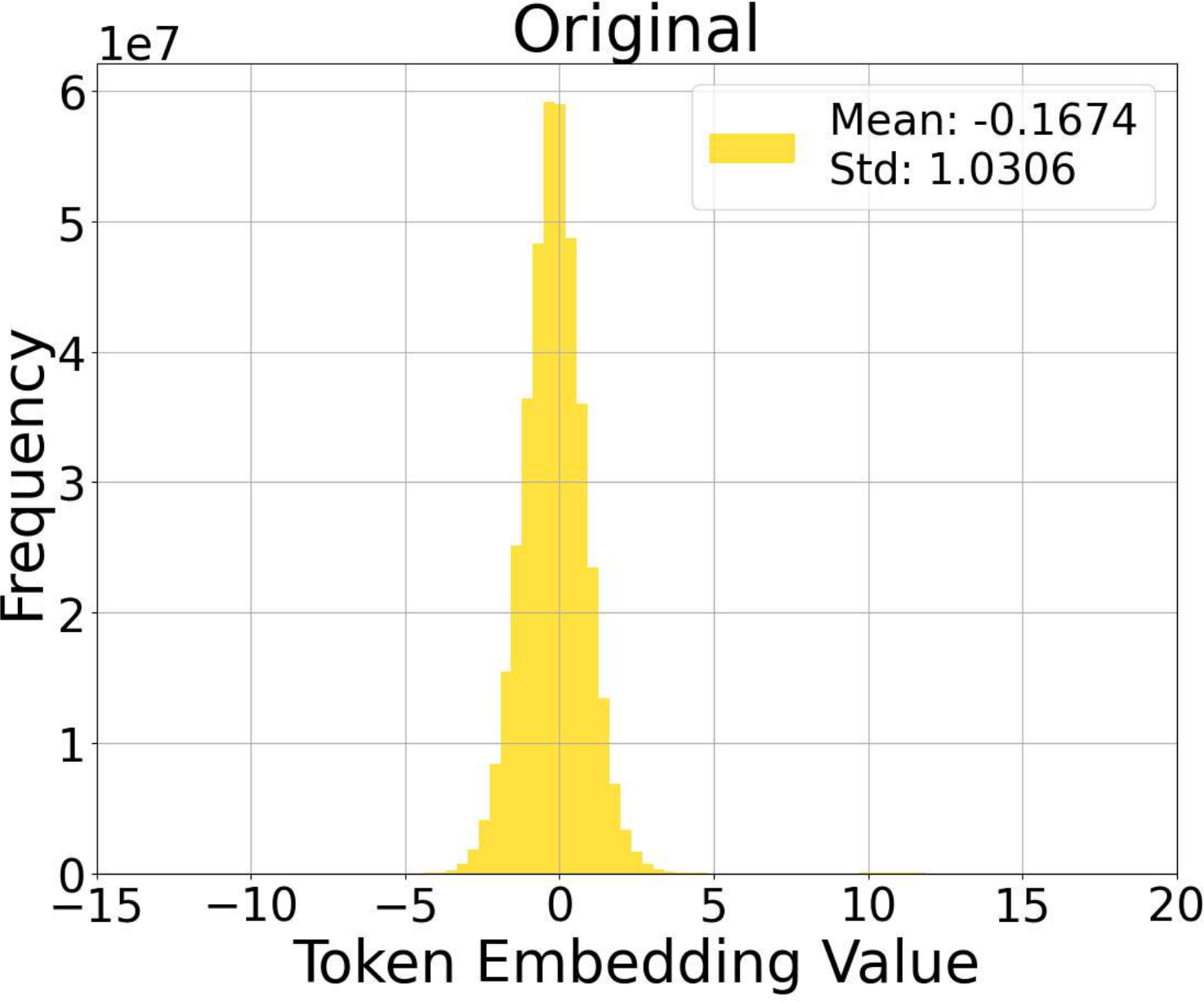}
    \put(2,3){\small (a)}  
  \end{overpic}
\end{minipage}

\vspace{0.3cm}

\begin{minipage}{0.35\textwidth}
  \centering
  \begin{overpic}[width=\linewidth]{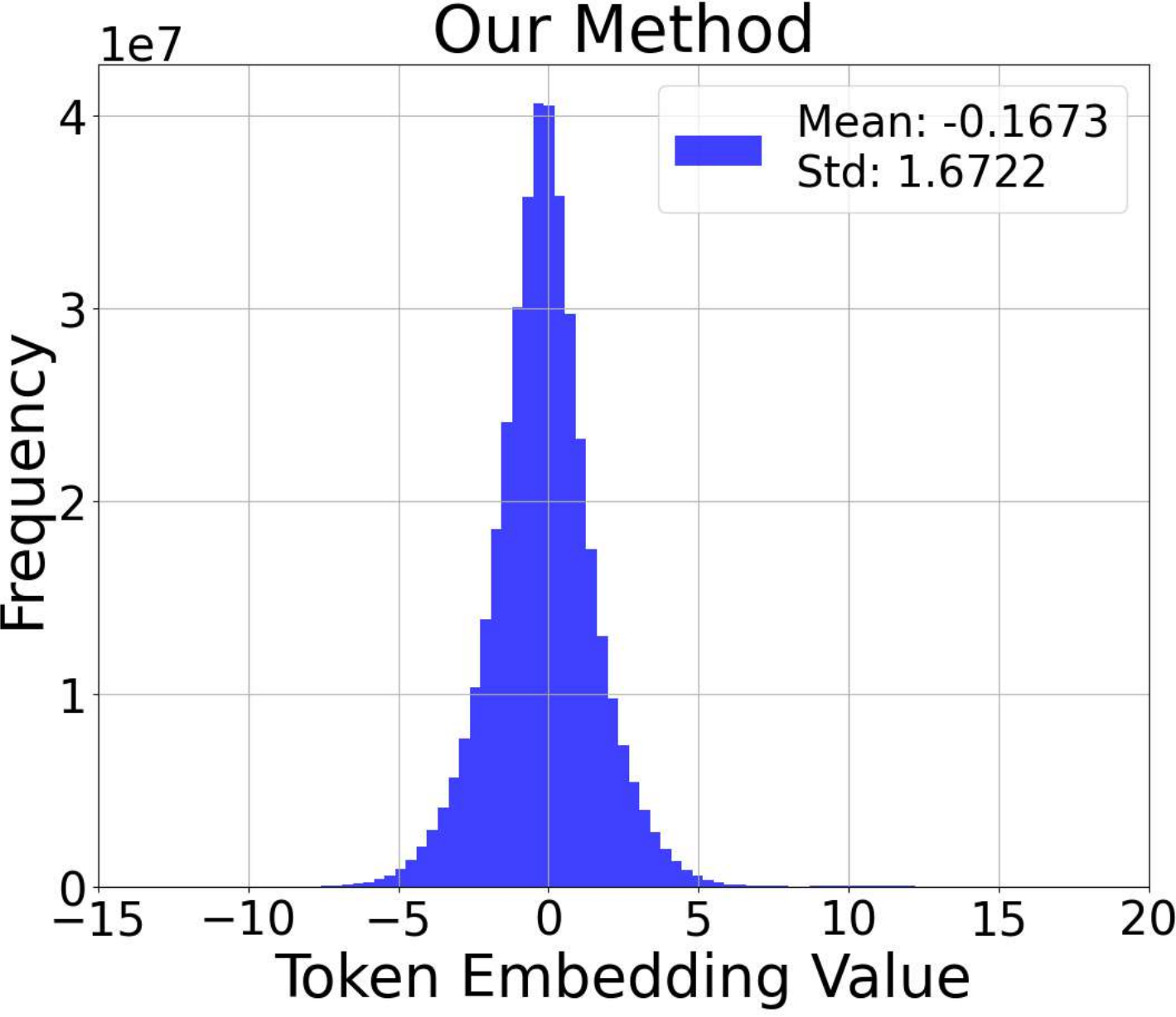}
    \put(2,3){\small (b)} 
  \end{overpic}
\end{minipage}

\vspace{0.3cm}

\begin{minipage}{0.35\textwidth}
  \centering
  \begin{overpic}[width=\linewidth]{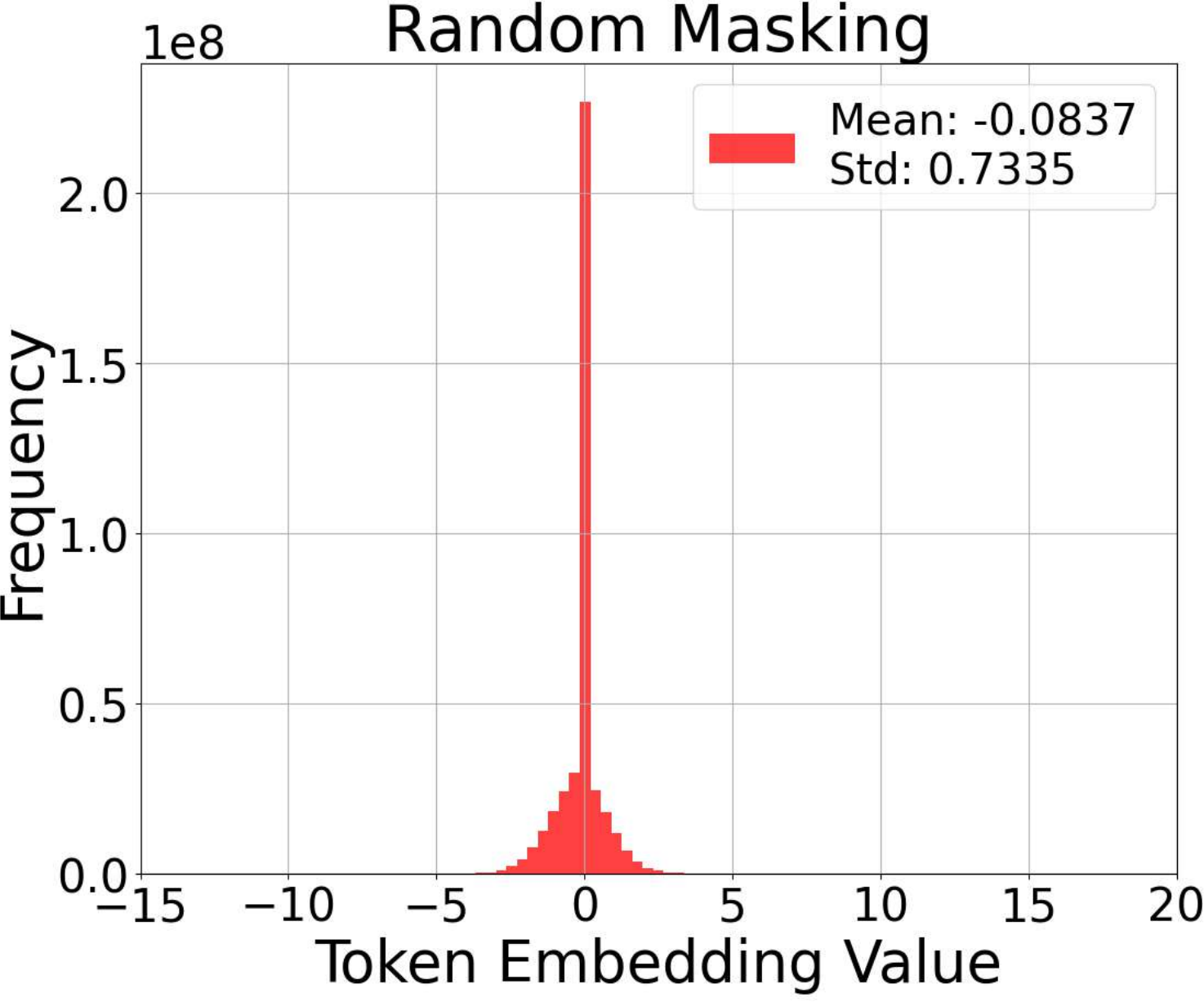}
    \put(2,3){\small (c)}  
  \end{overpic}
\end{minipage}

\vspace{-0.05cm}  

\caption{
        Comparison of token embedding distributions. 
        (a) Original token embeddings with mean $-0.1674$ and standard deviation $1.0306$;
        (b) Token embeddings of FPAN with mean $-0.1673$ close to that of the original embeddings but with a larger standard deviation of $1.6722$; 
        (c) Token embeddings of RM with smaller magnitude mean of
        $-0.0837$ and smaller standard deviation of $0.7335$.
    }
    \label{fig:embedding_histograms}
\end{figure}


In Section~\ref{sec:RandomMaskingVsFPAN}, we empirically demonstrated that FPAN achieves a more favorable balance between generation quality and replication compared to RM. In this section, we provide a theoretical interpretation of the differences between the two approaches.

We compare the impact of FPAN and RM on the distribution of token embeddings. We assume that each token embedding, denoted as \( \tau \), is sampled from a distribution with mean \( \mu_{\tau} \) and variance \( \sigma_{\tau}^2 \).

For FPAN, we denote the token embedding after adding noise \( \boldsymbol{\xi} \) as \( \tau' \), where \( \boldsymbol{\xi} \) is independent of \( \tau \) and is sampled from the distribution described in Equation~\ref{equation_5}. The resulting \( \tau' \) follows a distribution with mean \( \mu_{\tau}' \) and variance \( \sigma_{\tau}'^2 \), both of which can be derived as follows,
\begin{equation}
\mu'_{\tau} = \mathbb{E}[\tau + \boldsymbol{\xi}] = \mathbb{E}[\tau] + \mathbb{E}[\boldsymbol{\xi}] = \mu_{\tau} + 0 = \mu_{\tau}
\end{equation}
\begin{equation}
\begin{aligned}
\sigma'^2_{\tau} &= \text{Var}(\tau + \boldsymbol{\xi}) = \text{Var}(\tau) + \text{Var}(\boldsymbol{\xi}) \\
                 &= \sigma_{\tau}^2 + P \cdot W^2
\end{aligned}
\end{equation}

From the derivation above, it is evident that our method preserves the original mean of the token embedding distribution. According to~\cite{xu2022universal}, the mean of token embeddings serves as a strong representation of the overall sentence semantics. Thus, our method effectively retains the semantic content of the caption, which is crucial for maintaining the high quality of generated images. In addition, our method increases the variance of the token embedding distribution. This increased variance allows for greater diversity in token embeddings, thereby reducing the frequency with which the model encounters identical token embeddings during training, which in turn reduces the model’s tendency to memorize specific token information.

For RM, we denote the token embedding processed by RM as \( \tau'' \). The embedding \( \tau'' \) is computed as follows,
\begin{equation}
    \tau'' = m \cdot \tau, \quad m \sim \text{Bernoulli}(1-Q),
\end{equation}
where \( m \) is a Bernoulli random variable that equals 0 with probability \( Q \). We denote the mean and variance of the distribution that \( \tau'' \) follows as \( \mu_{\tau}'' \) and \( \sigma_{\tau}''^2 \), respectively. The values of \( \mu_{\tau}'' \) and \( \sigma_{\tau}''^2 \) can be derived as follows:
\begin{equation}
    \mu_{\tau}'' = \mathbb{E}[m \cdot \tau] = \mathbb{E}[m] \mathbb{E}[\tau] = (1-Q) \mu_{\tau}
\end{equation}
\begin{align}
\sigma_{\tau}''^2 &= \mathbb{E}[(\tau'')^2] - (\mathbb{E}[\tau''])^2 \notag \\
                  &= \mathbb{E}[m^2 \cdot \tau^2] - (1-Q)^2 \mu_{\tau}^2 \notag \\
                  &= (1-Q)(\sigma_{\tau}^2 + \mu_{\tau}^2) - (1-Q)^2 \mu_{\tau}^2 \notag \\
                  &= (1-Q) \sigma_{\tau}^2 + Q \mu_{\tau}^2 (1 - Q).
\end{align}

From the derivation results, we observe that because \( 1-Q < 1 \), that the mean of the tokens after random masking \( \mu_{\tau}'' \) is smaller in magnitude than the original \( \mu_{\tau} \). Furthermore, given that \( 1-Q < 1 \) and \( \mu_{\tau} \) is close to zero, it follows that \(( 1-Q) \sigma_{\tau}^2 < \sigma_{\tau}^2 \), and the term \( Q \mu_{\tau}^2 (1 - Q) \) approaches zero. As a result, the total variance \( \sigma_{\tau}''^2 = (1-Q) \sigma_{\tau}^2 + Q \mu_{\tau}^2 (1 - Q) \) is less than \( \sigma_{\tau}'^2 = \sigma_{\tau}^2 + P \cdot W^2 \). We conclude that RM alters the original semantic information and yields a smaller increase in variance than our method. 
These derivations suggest that RM is less effective than our approach in balancing the trade-off between generation quality and replication, possibly because it alters the mean of the token embedding distribution and fails to effectively increase the variance.

To experimentally show the differences of FPAN over RM, we analyze the impact of both methods on captions in the fine-tuning set. Specifically, we randomly select 5000 captions from the fine-tuning set, use \( W = 1.7 \) and \( P = 0.6 \) as the hyperparameters of FPAN and \( Q = 0.5 \) for RM. Figure~\ref{fig:embedding_histograms} presents the resulting means and standard deviations of the token embeddings after noise injection using each method. It can be observed that the mean of the token embeddings processed by our method remains nearly unchanged from the original token embeddings. In contrast, the magnitude of the mean of the embeddings under RM is approximately 50\% of the original value. Moreover, the standard deviation of the token embeddings produced by our method increases by 62.25\% relative to that of the original token embeddings, whereas the standard deviation resulting from RM decreases by 28.82\%. These results empirically validate our theoretical analysis.


\end{document}